\pdfoutput=1

\documentclass[11pt]{article}

\usepackage[]{acl2023}
\usepackage{amsmath}
\DeclareMathOperator*{\argmax}{arg\,max}
\usepackage{graphicx}
\usepackage{arydshln}
\usepackage{booktabs}
\usepackage{xcolor}
\usepackage{times}
\usepackage{latexsym}
\usepackage{algorithm}
\usepackage{algpseudocode}
\usepackage{float}

\usepackage[T1]{fontenc}

\usepackage[utf8]{inputenc}

\usepackage{microtype}

\usepackage{inconsolata}

\definecolor{model1}{RGB}{240, 29, 5}
\definecolor{model2}{RGB}{5, 95, 240}

\title{Co-training for Low Resource Scientific Natural Language Inference}

\author{Mobashir Sadat \mbox{   }\mbox{   }\mbox{   }\mbox{   } Cornelia Caragea\\
  Computer Science \\
  University of Illinois Chicago \\
  {\color{blue}\texttt{msadat3@uic.edu \mbox{    }  cornelia@uic.edu}} 
  }

\begin{document}
\maketitle
\begin{abstract}

Scientific Natural Language Inference (NLI) is the task of predicting the semantic relation between a pair of sentences extracted from research articles. The automatic annotation method based on distant supervision for the training set of \textsc{SciNLI} \cite{sadat-caragea-2022-scinli}, the first and most popular dataset for this task,
results in label noise which inevitably degenerates the performance of classifiers. In this paper, we propose a novel co-training method that assigns weights based on the training dynamics of the classifiers to the distantly supervised labels, reflective of the manner they are used in the subsequent training epochs. That is, unlike the existing semi-supervised learning (SSL) approaches, we consider the historical behavior of the classifiers to evaluate the quality of the automatically annotated labels. Furthermore, by assigning importance weights instead of filtering out examples based on an arbitrary threshold on the predicted confidence, we maximize the usage of automatically labeled data, while ensuring that the noisy labels have a minimal impact on model training. The proposed method obtains an improvement of ${\bf 1.5\%}$ in Macro F1 over the distant supervision baseline, and substantial improvements over several other strong SSL baselines. We make our code and data available on Github.\footnote{\url{https://github.com/msadat3/weighted_cotraining}}

\end{abstract}

\section{Introduction}
\vspace{-2mm}

Scientific Natural Language Inference (NLI) aims at predicting the semantic relation between a pair of sentences extracted from research articles. This task was recently proposed by \citet{sadat-caragea-2022-scinli} as a Natural Language Understanding (NLU) benchmark for scientific text along with a new dataset called \textsc{SciNLI}. The training set of \textsc{SciNLI} was constructed automatically using a distant supervision method \cite{mintz-etal-2009-distant}. The labels assigned using this method contain strong signals, and achieved a Macro F1 of $\sim78\%$ on the human annotated test set using pre-trained language models. Nevertheless, despite the strong signals, the labels assigned based on distant supervision still contain noise that inevitably degenerates the performance of classifiers. The noisy labels can be particularly harmful for training large scale deep learning models because these models can achieve zero training loss even for mislabeled examples simply by memorizing them \cite{zhang2021understanding, arpit2017closer}.

\looseness=-1
\vspace{-1mm}
Semi-supervised learning (SSL) methods such as pseudo-labeling, i.e., self-training  \cite{xie2020self, chen2021simple}, consistency regularization \cite{sajjadi2016regularization, sohn2020fixmatch}, and co-training \cite{blum1998combining} have emerged as promising approaches that utilize a small amount of human annotated data, and learn from a large amount of unlabeled data. Generally, these methods first train models with the limited human labeled data, and combine their most confident predictions (selected by applying a fixed high threshold, e.g., $0.9$ on confidence) for unlabeled examples with the human labeled data for subsequent training. While a fixed high threshold ensures high quality pseudo-labels, it also ignores a large fraction of diverse examples that have correct pseudo-labels but with lower confidence. Consequently, several dynamic thresholding mechanisms \cite{zhang2021flexmatch, xu2021dash} have been proposed to ensure a higher utilization of the pseudo-labels. However, a dynamic threshold still discards a large number of examples, compromises the pseudo-label quality, and is susceptible to error accumulation.

\vspace{-1mm}
In this paper, we propose a novel co-training approach for scientific NLI that takes the signal from labels assigned based on distant supervision, but their quality is discerned by two classifiers simultaneously trained on different regions of the data map \cite{swayamdipta2020dataset} that guide each other in the learning process. Unlike existing co-training approaches which also train two classifiers simultaneously in a cross-labeling manner by exchanging the most confident pseudo-labels, in our approach, the quality of each distantly supervised label is determined mutually by each classifier in the form of \textit{importance weights}, which are exchanged between the two classifiers. Thus, instead of exchanging the pseudo-labels, we exchange the classifiers' beliefs in the quality of the distantly supervised labels. In addition, we do not discard any examples but utilize all of them with different \textit{importance weights} to ensure that noisy labels have a minimal impact on model training. 

The weights in our approach are estimated by monitoring the training dynamics to capture the behavior of the classifiers on each example---\textit{easy}, \textit{hard}, and \textit{ambiguous}. Specifically, we calculate the weights based on both average \textbf{confidence}, and \textbf{variability} of the probability predicted over the training epochs for each label assigned based on distant supervision. Our weight calculation strategy ensures a high weight of the \textit{easy} examples which are more likely to be clean, and a low weight of the \textit{hard} examples which are more likely to be noisy for both classifiers; and a contrasting weight of the \textit{ambiguous} examples by each classifier to encourage divergence between them.

We explore our proposed approach by using a small amount of human annotated examples, and a large number of automatically annotated examples based on distant supervision from the \textsc{SciNLI} training set. Our experiments show that the proposed approach improves the performance by more than $\textbf{1.5\%}$ over distant supervision (i.e., models trained directly on the \textsc{SciNLI} training set); and obtains substantial improvements over co-training, co-teaching {\cite{han2018co}, and several other baselines designed for low-resource settings. 
Our key contributions can be summarized as follows:

\begin{itemize}
    \vspace{-2mm}
    \looseness=-1
    \item We develop a novel co-training approach for scientific NLI that utilizes the historical training dynamics to evaluate the quality of distantly supervised labels to assign \textit{importance weights} for these labels. 

     \item We thoroughly evaluate our proposed method by comparing its performance with distant supervision, co-training, co-teaching, and several other strong SSL methods.

    \item We present the first ever \textit{human annotated} training set for scientific NLI containing $2,000$ examples which we will make publicly available for future research.

\end{itemize}

\section{Related Work}

\paragraph{Scientific NLI} \citet{sadat-caragea-2022-scinli} proposed a distant supervision method based on linking phrases to automatically annotate large scale training datasets for the scientific NLI task and introduced the first dataset for this task which is called \textsc{SciNLI}. This task consists of four classes: \textsc{entailment}, \textsc{reasoning}, \textsc{contrasting} and \textsc{neutral}. The training examples for the former three classes are annotated automatically based on linking phrases indicative of these relations. For example, if the second sentence in an adjacent sentence pair starts with ``Therefore,'' the pair is labeled as \textsc{reasoning}. Random non-adjacent sentences are paired together and labeled as \textsc{neutral}. Using the same distant supervision method, \textsc{MSciNLI} \cite{sadat-caragea-2024-mscinli} was recently introduced to cover multiple scientific domains. The automatic annotation method results in noisy labels that can harm the performance and generalization of the classifiers.

\paragraph{Semi-supervised Learning} Self-training \cite{xie2020self, becker-etal-2013-avaya, NEURIPS2020_f23d125d, sadat-caragea-2022-learning} incorporates unlabeled data into model training by using three general steps: \textbf{a)} training a model using the available labeled examples; \textbf{b)} assigning pseudo-labels to unlabeled examples based on the model's prediction; \textbf{c)} selecting a subset of pseudo-labeled examples based on some quality assurance measures and using them for further model training in addition to the labeled examples. Consequently, the model gets exposed to more data which results in a better performance.

Co-training methods \cite{blum1998combining, wan2009co, 10.1145/2488388.2488430, chen2011automatic, qiao2018deep, zou-caragea-2023-jointmatch} employ a similar approach as self-training to incorporate pseudo-labeled data into model training with a key difference. Instead of training a single classifier as self-training, two classifiers are trained simultaneously which exchange their most confident pseudo-labels for further training. In contrast to the existing approaches, we do not discard any examples, and enable co-training by exchanging \textit{importance weights} indicative of the distantly supervised label quality, calculated based on the classifiers' training dynamics.  

Consistency regularization \cite{sajjadi2016regularization, temporalSSL} uses a consistency loss term (in addition to the supervised loss) that minimizes the distance between the predictions made by the model for different versions of the same unlabeled data. For example, FixMatch \cite{sohn2020fixmatch} first generates pseudo-labels for weak augmentations (e.g., replacing a random subset of tokens in a sentence with their synonyms) of unlabeled examples. A subset of these examples are then selected based on a fixed confidence threshold. FixMatch then uses strong augmentations (e.g., automatically translating a sentence to a different language and translating it back to the original language to get its paraphrased version) of the selected unlabeled examples as the input with these pseudo-labels as their target. Recently, there have been a plethora of consistency regularization based methods that build on top of FixMatch. For example, FlexMatch \cite{zhang2021flexmatch} uses a label-wise dynamic thresholding method based on the learning status for each label. SoftMatch \cite{chen2023softmatch} assigns soft weights to pseudo-labels based on the model confidence instead of filtering them out. Although SoftMatch does not filter out any examples, the soft weights are assigned based only on the current training iteration. That is, similar to other SSL approaches, \textsc{SoftMatch} ignores the historical behavior of the classifier on the unlabeled examples.

\vspace{-2mm}
\paragraph{Co-teaching}
The methods for learning from noisy labels generally rely on the classifier prediction during training to evaluate the quality of the labels. From this point of view, this research area is closely related to SSL. Co-teaching \cite{han2018co} is a method for training two classifiers simultaneously (similar to co-training) which exchange small-loss training instances (instances with low training loss) for further training. One of the key motivations behind training two classifiers is that each classifier will filter our different types of noise by learning complementary information. If the classifiers reach a consensus, they are no longer able to complement each other, reducing the learning process to single classifier training. Therefore, a divergence between the classifiers throughout the training process is crucial. To this end, De-coupling \cite{malach2017decoupling} and Co-teaching+ \cite{yu2019does} use only the examples for which the classifiers disagree in their predictions. However, similar to SSL approaches, these methods also evaluate the label quality based only on the current classifiers' predictions; ignore the history based on their training dynamics; and discard a large number of examples.

\section{Background}
We design our proposed approach based on our observations from the data map \cite{swayamdipta2020dataset} of \textsc{SciNLI}. In this section, we define the metrics used to characterize each example; plot them on a data map based on their characterizations; and discuss our observations from the data map that guide the development of our approach.

\label{sec:background_motivation}
\paragraph{Notations}

Consider a dataset $D = \{D^{l} \cup D^{a}\}$ where $D^{l} = \{({\bf x}^{i}, y^{i})\}_{i=1,...,n}$ is a small human labeled training set of size $n$; and $D^{a} = \{({\bf x}^{i}, \tilde{y}^{i})\}_{i=1,...,m}$ is an automatically labeled set of size $m$, constructed using distant supervision. ${\bf x}^{i}$ is a premise-hypothesis pair, $y^{i}$ is a human annotated label, $\tilde{y}^{i}$ is an automatically assigned label, and $n \ll m$. We denote $p (y|{\bf x;\theta)}$ as the predicted probability for an assigned label $y$ by a classifier $\theta$, given the premise-hypothesis pair ${\bf x}$.

\paragraph{Dataset Cartography} Dataset cartography \cite{swayamdipta2020dataset} is a tool for mapping and diagnosing a dataset by analyzing the training dynamics i.e., the model's behavior on each example during training. Particularly, each example is characterized by three metrics---\textbf{confidence}, \textbf{variability}, and \textbf{correctness}, defined below.

For an example $({\bf x},y)$, \citet{swayamdipta2020dataset} defines the confidence as the mean of the probabilities predicted for its assigned label $y$ over the training epochs \{$1,..,e$\}. That is, the confidence is calculated as follows:
\begin{equation}
\vspace{-1mm}
\label{eqn:confidence_calculation}
    c_{\theta}({\bf x},y) = \frac{1}{e}\sum_{t=1}^{e}p(y | {\bf x}; \theta^{t})
\end{equation}
Here, $c_{\theta}$ is a function that calculates the confidence of an example ({\bf x},y) by a classifier $\theta$.

The variability is defined as the standard deviation of the predicted probability for the assigned label $y$ over the training epochs. Specifically, variability is calculated as follows:
\vspace{-1mm}
\begin{equation}
\vspace{-2mm}
\label{eqn:variability_calculation}
    v_{\theta}({\bf x},y) = \sqrt{\frac{\sum^{e}_{t=1}(p(y | {\bf x}; \theta^{t}) - c_{\theta}({\bf x},y))^{2}}{e}}
    \vspace{-1mm}
\end{equation}

\vspace{1mm}
The correctness is calculated as the fraction of epochs in which the classifier predicts the assigned label $y$ correctly.

\begin{figure}[t]
\centering
    \includegraphics[scale=0.27]{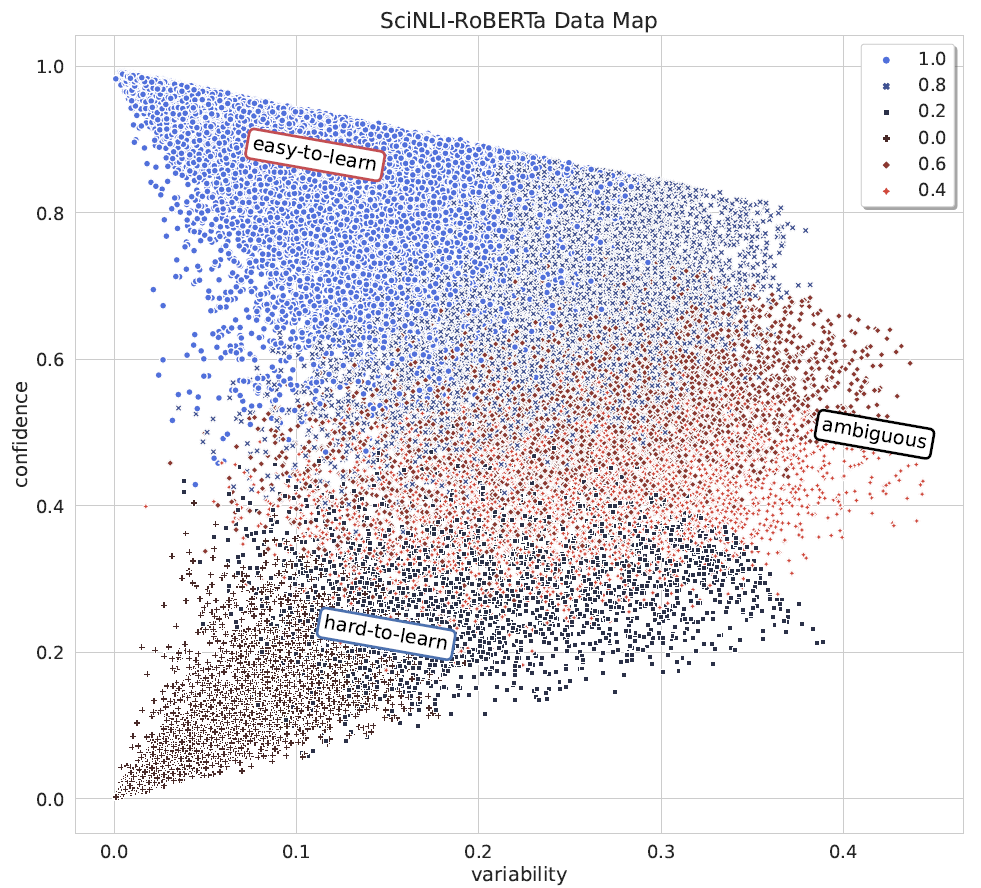}
    \vspace{-2mm}
    \caption{Data cartography of \textsc{SciNLI}. The colors and shapes indicate the correctness of each example.}
    \vspace{-4mm}
    \label{fig:SciNLI_cartography}
\end{figure}

We train a \textsc{RoBERTa} \cite{liu2019roberta} classifier on the full \textsc{SciNLI} training set $D$, and plot $55$K randomly selected examples in Figure \ref{fig:SciNLI_cartography} based on their training dynamics. Note that, we plot only $55$K examples (instead of all $101$K) for clarity.

\looseness=-1
The plot shows that the examples with high confidence and low variability (top left corner) have a high correctness. That is, the model's prediction consistently matches with the assigned label for these examples. Therefore, they are \textit{easy} to learn for the model, and the assigned labels are more likely to be correct. In contrast, the examples with low confidence and low variability (bottom left corner of the plot) have a low correctness. In other words, the model's predictions disagree with the assigned labels consistently. Thus, the examples in this region are \textit{hard} for the model to learn, and the assigned labels are more likely to be incorrect. The examples with a high variability (right side of the plot) shows a moderate correctness, and confidence. This indicates the model is uncertain about these examples, and their predictions occasionally match with the assigned labels. In our approach, we employ a weighting strategy that utilizes the information from the data maps to assign importance weights to distantly supervised examples during training.

\section{Proposed Approach}
\label{section:proposed_approach}

\looseness=-1
We now present our co-training approach where we train two classifiers simultaneously that capture complementary information from the data. Unlike the existing co-training approaches, we utilize all examples in dataset $D^{a}$, and do not filter out anything. Rather, our approach assigns importance weights to the automatically annotated examples. Based on these weights, we decide the impact of each automatically annotated example in subsequent training of the classifiers. We devise the weights of the automatically labeled examples guided by the training dynamics of the classifiers in the form of confidence and variability of each example over the training epochs. 
Furthermore, in contrast to the existing approaches which exchange pseudo-labels between classifiers, we exchange the weights calculated based on the historical behavior of the classifiers in evaluating the quality of each example. 
We ensure that the two classifiers learn complementary information by employing a weight calculation strategy that maintains their divergence. An overview of our approach can be seen in Alg. \ref{alg:weighted_co_training}.

\subsection{Weight assignment} 
The two classifiers that we co-train are denoted as $\theta_{\textcolor{model1}{1}}$ and $\theta_{\textcolor{model2}{2}}$. We now describe how we assign importance weights to the automatically annotated examples at each epoch to train the classifiers. 

We calculate two sets of weights in our co-training method based on the training dynamics of the classifiers on the distantly supervised labels. Specifically, for each $\tilde{y}^{i}$ in $D^a$, we calculate $c_{\theta_{\textcolor{model1}{1}}}({\bf x^{i}}, \tilde{y}^{i})$, and  $c_{\theta_{\textcolor{model2}{2}}}({\bf x^{i}}, \tilde{y}^{i})$ using Eq. \ref{eqn:confidence_calculation}; and  $v_{\theta_{\textcolor{model1}{1}}}({\bf x^{i}}, \tilde{y}^{i})$, and  $v_{\theta_{\textcolor{model2}{2}}}({\bf x^{i}}, \tilde{y}^{i})$ using Eq. \ref{eqn:variability_calculation}. The weights $\lambda_{\textcolor{model1}{1}}^{i}$, and $\lambda_{\textcolor{model2}{2}}^{i}$ for each example to be used for subsequent training of $\theta_{\textcolor{model2}{2}}$, and $\theta_{\textcolor{model1}{1}}$, respectively are calculated as follows:
\vspace{-2mm}
\begin{equation}
\label{equation:weight_assignment_1}
    \lambda^{i}_{\textcolor{model1}{1}} = c_{\theta_{\textcolor{model1}{1}}}({\bf x^{i}}, \tilde{y}^{i}) + v_{\theta_{\textcolor{model1}{1}}}({\bf x^{i}}, \tilde{y}^{i})
\end{equation}
\vspace{-5mm}
\begin{equation}
\label{equation:weight_assignment_2}
    \lambda_{\textcolor{model2}{2}}^{i} =c_{\theta_{\textcolor{model2}{2}}}({\bf x^{i}}, \tilde{y}^{i}) - v_{\theta_{\textcolor{model2}{2}}}({\bf x^{i}}, \tilde{y}^{i})
\end{equation}
Therefore, the weights for the \textit{easy} examples (high confidence, low variability) will be high for both classifiers, and the weights for the \textit{hard} (low confidence, low variability) examples will be low for both classifiers. For the \textit{ambiguous} examples where the variability is high, the weight for $\theta_{\textcolor{model1}{1}}$ (i.e, $\lambda^{i}_{\textcolor{model2}{2}}$) will be low given that the classifiers show a moderate confidence for these examples (see the right side of Figure \ref{fig:SciNLI_cartography}). For example, a confidence of $0.4$, and a variability of $0.3$ results in low weight of $0.1$. On the other hand, if the variability is high, the weight for $\theta_{\textcolor{model2}{2}}$ (i.e, $\lambda^{i}_{\textcolor{model1}{1}}$) will remain high even if the confidence is moderate. For example, a confidence of $0.4$, and a variability of $0.3$ results in high weight of $0.7$. We normalize the weights using a min-max normalization method. Therefore, both sets of weights are scaled to a range of $0$ to $1$.

\algnewcommand\algorithmicforeach{\textbf{for each}}
\algdef{S}[FOR]{ForEach}[1]{\algorithmicforeach\ #1\ \algorithmicdo}

\algrenewcommand\algorithmicensure{\textbf{Step}}

\begin{algorithm*}[t]

\caption{Weighted Co-training for Scientific NLI}\label{alg:weighted_co_training}
\begin{algorithmic}[1]
\small
\Require Human labeled set, $D^{l} = \{({\bf x}^{i}, y^{i})\}_{i=1,...,n}$; automatically labeled set, $D^{a} = \{({\bf x}^{i}, \tilde{y}^{i})\}_{i=1,...,m}$; Maximum training epochs $E$; learning rate $\eta$.

\State Train $\theta_{\textcolor{model1}{1}}$, and  $\theta_{\textcolor{model2}{2}}$ on two randomly divided equal subsets of $D^l$; for each automatically annotated example $i$ in $D^a$, store the probability $p (\tilde{y^{i}}|{\bf x^{i};\theta_{\textcolor{model1}{1}})}$ and $p (\tilde{y}^{i}|{\bf x^{i};\theta_{\textcolor{model2}{2}}})$ after each epoch.

\ForEach {$(x^{i}, \tilde{y}^{i}) \in D^{a}$ }
\State Calculate confidences $c_{\theta_{\textcolor{model1}{1}}}({\bf x^{i}}, \tilde{y}^{i})$ and $c_{\theta_{\textcolor{model2}{2}}}({\bf x^{i}}, \tilde{y}^{i})$, and variabilities $v_{\theta_{\textcolor{model1}{1}}}({\bf x^{i}}, \tilde{y}^{i})$, and $v_{\theta_{\textcolor{model2}{2}}}({\bf x^{i}}, \tilde{y}^{i})$ from the stored probabilities using Eq. \ref{eqn:confidence_calculation} and Eq. \ref{eqn:variability_calculation}.

\State $\lambda^{i}_{\textcolor{model1}{1}} \gets c_{\theta_{\textcolor{model1}{1}}}({\bf x^{i}}, \tilde{y}^{i}) + v_{\theta_{\textcolor{model1}{1}}}({\bf x^{i}}, \tilde{y}^{i})$; and $\lambda^{i}_{\textcolor{model2}{2}} \gets c_{\theta_{\textcolor{model2}{2}}}({\bf x^{i}}, \tilde{y}^{i}) - v_{\theta_{\textcolor{model2}{2}}}({\bf x^{i}}, \tilde{y}^{i})$. \Comment{\textcolor{gray!50}{Assign initial importance weights.}}
\EndFor

\State Re-initialize $\theta_{\textcolor{model1}{1}}$, $\theta_{\textcolor{model2}{2}}$ \Comment{\textcolor{gray!50}{Re-initialize the classifiers for co-training.}}

\For{$e=1$ to $E$}

\ForEach {mini-batch $B \in D^{a}$ }

\State $\mathcal{L}_{\textcolor{model1}{1}} \gets \frac{1}{|B|} \sum_{i=0}^{|B|} \lambda^{i}_{\textcolor{model2}{2}} * H(\tilde{y}^{i}, p_{d}({\bf x}^{i}; \theta_{\textcolor{model1}{1}}))$;  $\mathcal{L}_{\textcolor{model2}{2}} \gets \frac{1}{|B|} \sum_{i=0}^{|B|} \lambda^{i}_{\textcolor{model1}{1}} * H(\tilde{y}^{i}, p_{d}({\bf x}^{i}; \theta_{\textcolor{model2}{2}}))$. 
\Comment{\textcolor{gray!50}{Calculate cross-entropy loss.}} 

\ForEach {\textsc{DS-labeled} example $i \in B$ }
\State Update $c_{\theta_{\textcolor{model1}{1}}}({\bf x^{i}}, \tilde{y}^{i})$ and $c_{\theta_{\textcolor{model2}{2}}}({\bf x^{i}}, \tilde{y}^{i})$ using Eq. \ref{eqn:confidence_calculation};  update $v_{\theta_{\textcolor{model1}{1}}}({\bf x^{i}}, \tilde{y}^{i})$ and $v_{\theta_{\textcolor{model2}{2}}}({\bf x^{i}}, \tilde{y}^{i})$ using Eq \ref{eqn:variability_calculation}.

\State $\lambda^{i}_{\textcolor{model1}{1}} \gets c_{\theta_{\textcolor{model1}{1}}}({\bf x^{i}}, \tilde{y}^{i}) + v_{\theta_{\textcolor{model1}{1}}}({\bf x^{i}}, \tilde{y}^{i})$; and $\lambda^{i}_{\textcolor{model2}{2}} \gets c_{\theta_{\textcolor{model2}{2}}}({\bf x^{i}}, \tilde{y}^{i}) - v_{\theta_{\textcolor{model2}{2}}}({\bf x^{i}}, \tilde{y}^{i})$. \Comment{\textcolor{gray!50}{Update the weights.}}
\EndFor

\State $\theta_{\textcolor{model1}{1}} \gets \theta_{\textcolor{model1}{1}} - \eta * \nabla \mathcal{L}_{\textcolor{model1}{1}}$; $\theta_{\textcolor{model2}{2}} \gets \theta_{\textcolor{model2}{2}} - \eta * \nabla \mathcal{L}_{\textcolor{model2}{2}}$  \Comment{\textcolor{gray!50}{Update the parameters of $\theta_{1}$, and $\theta_{1}$}}

\EndFor

\EndFor

\State Fine-tune $\theta_{\textcolor{model1}{1}}$, and $\theta_{\textcolor{model2}{2}}$ on two randomly divided equal subsets of human annotated data with a learning rate lower than $\eta$.

\vspace{-1mm}
\end{algorithmic}
\end{algorithm*}

Our weighting strategy ensures that a) the \textit{easy} examples which are likely to be correctly labeled, participate in the training of both classifiers with high weights; 
b) the \textit{hard} examples which are likely to be incorrectly labeled, participate in the training of both classifiers with low weights; 
and c) the \textit{ambiguous} examples for which the classifiers are uncertain about the assigned labels, participate in the training of one classifier with high weights, and participate in the training of the other classifier with low weights. This allows the classifiers to remain divergent, and ensure that they capture complementary information from the data. That is, in contrast to prior methods (e.g., De-coupling and Co-teaching+), we enforce divergence between the classifiers by employing a contrasting usage of the ambiguous examples for the classifiers.

\subsection{Weighted Co-training} 
\label{section:wct}
We co-train the two classifiers $\theta_{\textcolor{model1}{1}}$ and $\theta_{\textcolor{model2}{2}}$ in three major steps. They are described below.

\vspace{-2mm}
\paragraph{Step 1: Initial weight assignment.} 

To assign the initial weights to the automatically annotated examples in $D^a$, we train both classifiers $\theta_{\textcolor{model1}{1}}$ and $\theta_{\textcolor{model2}{2}}$ using two randomly divided subsets of equal size of the human annotated set in $D^l$. At the end of each epoch, we record the predicted probabilities $p (\tilde{y}^{i}|{\bf x^{i};\theta_{\textcolor{model1}{1}})}$ and $p (\tilde{y}^{i}|{\bf x^{i};\theta_{\textcolor{model2}{2}}})$ for each example in $D^a$. We then calculate the confidences, and variabilities based on the recorded probabilities using Eq. \ref{eqn:confidence_calculation} and \ref{eqn:variability_calculation}, respectively. Based on these confidences and variabilities, the initial weights for the automatically annotated examples 
are then calculated using Eq. \ref{equation:weight_assignment_1} and \ref{equation:weight_assignment_2}.

\vspace{-1mm}
\paragraph{Step 2: Co-training epochs.} 
We co-train $\theta_{\textcolor{model1}{1}}$ and $\theta_{\textcolor{model2}{2}}$ using $D^a$. We do not include $D^l$ in the co-training epochs because assigning importance weights to a combination of human labeled, and automatically labeled examples results in an unfair advantage to the former. Therefore, $D^l$ is utilized separately in an additional step.

Both classifiers are re-initialized before we start the co-training epochs. At each epoch, the cross-entropy loss for each classifier for an example is scaled by its assigned weight by the other classifier. In particular, for each mini-batch $B$, the losses $\mathcal{L}_{\textcolor{model1}{1}}$ and $\mathcal{L}_{\textcolor{model2}{2}}$ for $\theta_{\textcolor{model1}{1}}$ and $\theta_{\textcolor{model2}{2}}$ are calculated as follows:

\begin{equation}
\label{equation:loss_scaling_1}
    \mathcal{L}_{\textcolor{model1}{1}} = \frac{1}{|B|} \sum_{i=0}^{|B|} \lambda^{i}_{\textcolor{model2}{2}} * H(\tilde{y}^{i}, p_{d}({\bf x}^{i}; \theta_{\textcolor{model1}{1}}))
\end{equation}
\begin{equation}
\label{equation:loss_scaling_2}
    \mathcal{L}_{\textcolor{model2}{2}} = \frac{1}{|B|} \sum_{i=0}^{|B|} \lambda^{i}_{\textcolor{model1}{1}} * H(\tilde{y}^{i}, p_{d}({\bf x}^{i}; \theta_{\textcolor{model2}{2}}))
\end{equation}
Here, $p_{d}$ is the probability distribution over the labels predicted by a classifier for an automatically annotated example $i$ in $B$, and $H$ is the standard cross-entropy loss.

After calculating the loss for both classifiers, we update the confidences, and variabilities using Eq. \ref{eqn:confidence_calculation}, and \ref{eqn:variability_calculation}. Note that the probabilities predicted by the classifiers in the last epoch in Step 1, are used as initial probabilities for both classifiers, and are included in calculating the mean, and the standard deviation in the co-training epochs. 
We then update both sets of weights using Eq. \ref{equation:weight_assignment_1} and \ref{equation:weight_assignment_2}, based on the updated confidences and variabilities, to be used in the next epoch. Therefore, the classifiers co-train each other by exchanging the complementary information from their respective training dynamics over the epochs.    

\vspace{-1mm}
\paragraph{Step 3: Fine-tuning.} The co-training epochs continue until we reach a pre-defined maximum epoch $E$. We then fine-tune $\theta_{\textcolor{model1}{1}}$ and $\theta_{\textcolor{model2}{2}}$  using two randomly divided subsets of equal size of $D^l$. 

\vspace{-2mm}
\subsection{Classifier Ensembling} 
\label{sec:ensembling}

We ensemble the co-trained classifiers by performing an element-wise average of their softmax output (i.e., the probability distribution). 
We then take the $\argmax$ of the ensembled probability distribution to predict the labels for the test examples.

\section{Experiments} 
This section describes the baselines, our models, experimental settings, and results. 

\subsection{Baselines} 
The baselines that we consider for evaluating our proposed method are described below.

\paragraph{\textsc{Fully-supervised (FS)}} This baseline trains a fully supervised classifier using human labeled or \textsc{H-labeled} data. We explore two variants of this baseline --- \textsc{FS (H-labeled 1K)}, and \textsc{FS (H-labeled 2K)} which use $1,000$ randomly sampled class-balanced \textsc{H-labeled} examples, and all $2000$ \textsc{H-labeled} examples, respectively.

\vspace{-1mm}
\paragraph{\textsc{Distant Supervision (DS)}} This is the distantly supervised counterpart of the \textsc{FS} baselines. We experiment with three variants --- \textsc{DS-labeled (1K)}, \textsc{DS-labeled (2K)}, and \textsc{DS-labeled (101K)} using randomly sampled $1000$, $2000$, and all $101,412$ examples in the \textsc{SciNLI} training set, respectively.

\vspace{-1mm}
\paragraph{\textsc{Back-translation (BT)} \cite{yu2018fast}} A data augmentation method based on machine translation where each example is translated to French and then translated back to English to get their paraphrased versions. The paraphrased and original versions of the dataset are combined and models are trained on this larger set. We explore back-translation on top of both \textsc{FS} baselines.

\vspace{-1mm}
\paragraph{\textsc{DBST} \cite{sadat-caragea-2022-learning}} A self-training method for NLI where examples are selected based on whether the automatically assigned label (e.g., distantly supervised label in \textsc{SciNLI}) of each example matches with the predicted label by the classifier and whether the confidence for the predicted label is above a pre-defined threshold. 

\vspace{-1mm}
\paragraph{\textsc{FixMatch} \cite{sohn2020fixmatch}} A consistency regularization method based on strong and weak augmentations of unlabeled data. We use synonym replacement as the weak augmentation, and backtranslation as the strong augmentation technique. FixMatch does not consider the distantly supervised labels, and relies on the classifier prediction.  

\vspace{-1mm}
\paragraph{\textsc{FlexMatch} \cite{zhang2021flexmatch}} \textsc{FlexMatch} is an SSL approach that uses \textsc{FixMatch} as its backbone. In contrast to \textsc{FixMatch}, \textsc{FlexMatch} applies different confidence thresholds for different classes based on their learning status. 

\vspace{-1mm}
\paragraph{\textsc{SoftMatch} \cite{chen2023softmatch}} \textsc{SoftMatch} also uses \textsc{FixMatch} as its backbone method. However, instead of filtering out pseudo-labeled examples, it assigns importance weights based on the model confidence in the current training epoch. 

\vspace{-1mm}
\paragraph{\textsc{Co-training}} A vanilla \textsc{Co-training} approach which does not make use of distantly supervised labels, and exchanges a subset of the predictions selected based on a confidence threshold between the classifiers for training them. 

\vspace{-1mm}
\paragraph{\textsc{Co-teaching} \cite{han2018co}} \textsc{Co-teaching} is a method for learning from noisy labels. It does not use any human annotated data, and trains two classifiers simultaneously. The classifiers exchange the small-loss examples (i.e., examples with low cross-entropy loss) which are used for their training.   

\vspace{-1mm}
\paragraph{\textsc{Co-teaching}+ \cite{yu2019does}} This method is similar to \textsc{Co-teaching}, except it enforces a divergence between the classifiers by using only the small-loss examples for which the classifiers disagree in their predictions.

\setlength\dashlinedash{0.2pt}
\setlength\dashlinegap{1.5pt}
\setlength\arrayrulewidth{0.3pt}
\begin{table*}[t]
\centering
\small
  \begin{tabular}{ l c c c c }
    \toprule
&  \multicolumn{2}{c}{\bf SciBERT} & \multicolumn{2}{c}{\bf RoBERTa}\\\cmidrule(lr){2-3} \cmidrule(lr){4-5}
   {\bf Method} & {\bf Macro F1}     & {\bf Acc} & {\bf Macro F1}     & {\bf Acc} \\ 
   \midrule
   
    {\textsc{DS-labeled (1K)}} & $61.21\pm1.8$   & $61.35\pm1.7$ & $63.29\pm2.9$ & $63.45\pm2.6$ \\
     {\textsc{FS (H-labeled 1K)}} & $65.31\pm0.7$   & $65.39\pm0.6$ & $67.70\pm0.2$ & $67.78\pm0.2$ \\
     {\textsc{BT (H-labeled 1K)}} & $64.13\pm0.4$   & $64.17\pm0.3$ & $67.73\pm1.1$ & $67.78\pm1.1$ \\
     \hdashline
    
    {\textsc{DS-labeled (2K)}} & $64.74\pm1.6$   & $64.75\pm1.4$ & $67.01\pm1.6$ & $67.02\pm1.4$ \\
    {\textsc{FS (H-labeled 2K)}} & $67.46\pm1.1$   & $67.51\pm1.1$ & $70.13\pm0.3$ & $70.14\pm0.2$ \\
    {\textsc{BT (H-labeled 2K)}} & $68.12\pm1.9$   & $68.18\pm1.9$ & $70.23\pm0.7$ & $70.29\pm0.7$ \\
    \hdashline
    {\textsc{DS-labeled (101K)}}  & $77.53\pm0.5$   & $77.52\pm0.5$ & $78.08\pm0.4$ & $78.11\pm0.4$ \\
    \hdashline
    
    {\textsc{DBST} \cite{sadat-caragea-2022-learning}} & $73.61\pm0.4$   & $73.59\pm0.5$ & $73.74\pm0.3$ & $73.72\pm0.3$ \\

    \textsc{FixMatch} \cite{sohn2020fixmatch} & $68.21 \pm 0.8$ & $68.21 \pm 0.8$ & $71.27 \pm 0.2$ & $71.32 \pm 0.2$\\
    \textsc{FlexMatch} \cite{zhang2021flexmatch} & $68.60 \pm 0.9$ & $68.56 \pm 0.8$ & $71.64 \pm 0.3$ & $71.65 \pm 0.4$\\
    \textsc{SoftMatch} \cite{chen2023softmatch} & $68.77 \pm 1.3$ & $68.70 \pm 1.3$ & $71.80 \pm 0.2$ & $71.65 \pm 0.4$\\
    \hdashline
    \textsc{Co-training} & $68.29 \pm 1.7$ & $68.28 \pm 1.7$ & $70.35 \pm 0.2$ & $70.47 \pm 0.2$\\
    \textsc{Co-teaching} \cite{han2018co} & $78.06 \pm 0.3$ & $78.05 \pm 0.3$ & $78.72 \pm 0.4$ & $78.77 \pm 0.5$\\
    \textsc{Co-teaching+} \cite{yu2019does} & $76.27 \pm 0.3$ & $76.24 \pm 0.3$ & $76.47 \pm 0.4$ & $76.45 \pm 0.5$\\
    \hdashline
    
    \hdashline

    \hdashline
    {\textsc{WCT - cc}} & $78.35 \pm 0.5$ & $78.36 \pm 0.5$ & $79.12^{*} \pm 0.2$ & $79.15^{*} \pm 0.1$\\
    {\textsc{WCT - cv}} & $\textbf{78.55}^{*} \pm \textbf{0.4}$ & $\textbf{78.57}^{*} \pm \textbf{0.4}$ & $\textbf{79.62}^{*\#} \pm \textbf{0.3}$ & $\textbf{79.65}^{*\#} \pm \textbf{0.3}$\\

    \bottomrule
  \end{tabular}
  \caption{The Macro F1 and Accuracies of different methods on \textsc{SciNLI}. Best scores are in bold. * and \# indicate statistically significant improvements over \textsc{DS-labeled (101K)} and \textsc{Co-teaching} (our best performing baseline), respectively, according to a paired t-test with $p < 0.05$.} 
  
    \label{table:main_results}
\end{table*}

\subsection{Our Models}

We train the following models based on our proposed approach.
\vspace{-1mm}
\paragraph{\textbf{Weighted Co-training (WCT) - cv}}  \textsc{WCT - cv} denotes the classifiers we train using our proposed co-training approach described in Section \ref{section:proposed_approach}.
\vspace{-1mm}
\paragraph{\textbf{Weighted Co-training (WCT) - cc}} A variant of our proposed co-training approach where the weights for the automatically labeled examples are calculated only based on the confidences, and the variabilities are not used.

\subsection{Experimental Settings} 

\paragraph{Dataset}

We consider the full \textsc{SciNLI} training set containing $101$K examples as our dataset $D$. We then employ three expert annotators to manually curate $2,000$ examples from $D$ which are used as the human labeled set, $D^{l}$. A detailed description of our manual data annotation method is available in Appendix \ref{sec:annotation_details}. We then remove any example that was extracted from the same paper as the examples in $D^l$ from $D$. This resulted in $97$K examples which are used as $D^a$ in our approach.

\vspace{-2mm}
\paragraph{Implementation Details} We experiment with the base variants of both \textsc{SciBERT} \cite{beltagy-etal-2019-scibert} and \textsc{RoBERTa} \cite{liu2019roberta} for the baselines and our models. More implementation details are available in Appendix \ref{sec:implementation}. Each experiment is run three times with different random seeds. The average and standard deviations of the Macro F1 and accuracy of different methods are in Table \ref{table:main_results}. Our findings are described below.

\subsection{Results \& Observations}
\vspace{-1mm}

\paragraph{Weighted co-training vs. distant supervision and data augmentation.} As we can see from the results, \textsc{WCT - cv} shows significant improvements over the \textsc{DS-labeled (101K)} baseline. In addition, our proposed approach substantially outperforms all \textsc{FS}, and \textsc{BT} baselines. These improvements in performance illustrate that, our proposed co-training approach is successful in reducing the impact of noisy labels on classifier training, resulting in a better performance. 

\looseness=-1
\paragraph{Weighted co-training vs. co-teaching.} Comparing \textsc{WCT - cv} and \textsc{Co-teaching}, we can see that \textsc{WCT - cv} shows better performance. Furthermore, in contrast to our method, the improvement shown by \textsc{Co-teaching} over the \textsc{DS-labeled (101K)} is not statistically significant. We can also see that \textsc{Co-teaching+} which aims at maintaining the divergence between the classifiers by using only disagreement data for training the classifiers, shows a poor performance. These observations illustrate that a) using only the predictions from the current epoch is myopic, and valuable information from the prior epochs is lost; and b) while the disagreement data possibly improves the divergence of the classifiers, the agreement data also contains diverse patterns, and are necessary.   

\looseness=-1
\paragraph{\textsc{WCT - cv} vs. \textsc{WCT - cc}.} While \textsc{WCT - cc} shows a higher Macro F1 than all other baselines, it shows a lower performance compared with \textsc{WCT - cv}. Note that \textsc{WCT - cc} uses only confidence to assign the weights to the automatically labeled examples for both classifiers. That is, unlike \textsc{WCT - cv}, \textsc{WCT - cc} does not enforce a divergence between the classifiers by using contrasting weights for the \textit{ambiguous} examples. Therefore, our weighting strategy in \textsc{WCT - cv} indeed improves the divergence between the classifiers, which results in improved performance.

\paragraph{Weighted co-training  vs. vanilla co-training.} The results show that the \textsc{Co-training} baseline which does not make use of the distantly supervised labels, and uses the most confident classifier predictions, achieves a much lower Macro F1 than the \textsc{DS-labeled (101K)} baseline. In addition, it only shows marginal improvements over the \textsc{FS} baselines. Therefore, distantly supervised labels contain strong signals which can be beneficial for model training, rather than using model predictions which can be unreliable, especially in the initial training epochs. 

\vspace{-1mm}
\paragraph{Weighted co-training vs. other SSL approaches.}
We can see that the \textsc{SSL} approaches---\textsc{FixMatch}, \textsc{FlexMatch}, and \textsc{SoftMatch} show a much lower performance than \textsc{DS-labeled (101K)}. Note that these single-classifier (pseudo-labeling) approaches do not utilize the distantly supervised labels, and rely only on the classifier prediction similar to vanilla co-training. Given their susceptibility to error accumulation, and the challenging nature of the scientific NLI task, they fail to show any promising performance. \textsc{DBST} on the other hand utilizes the signal from distant supervision in addition to model prediction. Consequently, \textsc{DBST} shows improved performance over the other SSL baselines. However, it filters out examples based on a confidence threshold, resulting in a much lower performance compared to \textsc{WCT - cv}.

\section{Analysis}
\vspace{-2mm}
Our analysis consists of two parts. First, we perform various ablation experiments (\S \ref{sec:ablation}). 
Next, we analyze the generalizability of our co-training approach to other NLP tasks (\S \ref{sec:generalizability}). We also analyze the robustness of \textsc{WCT - cv} by evaluating its out-of-domain performance using \textsc{MSciNLI} \cite{sadat-caragea-2024-mscinli} in Appendix \ref{appendix:robustness}.

\setlength\dashlinedash{0.2pt}
\setlength\dashlinegap{1.5pt}
\setlength\arrayrulewidth{0.3pt}
\begin{table}[t]
\centering
\small
  \begin{tabular}{l c c }
    \toprule
{\bf Approach} & {\bf SciBERT}     & {\bf RoBERTa} \\ 
   \midrule

   \textsc{Simple FT - ensembled} & $78.03$ & $78.52$\\
   \textsc{WST - ensembled} & $78.46$ & $79.33$  \\

   \textsc{WST - R} & $78.25$ & $78.50$\\
   \textsc{WCT - cv both 2K}  & $78.29$ & $79.35$\\
   \textsc{WCT - cvh} & $78.31$ & $79.29$\\
   \textsc{WCT - cv} & $\textbf{78.55}$ & $\textbf{79.62}^{*\#}$ \\

    \bottomrule
  \end{tabular}
  \vspace{-2mm}
  \caption{Macro F1 scores from the ablation experiments compared with \textsc{WCT - cv}. * and \# indicate statistically significant improvements over \textsc{Simple FT - ensembled} and \textsc{WST - R}, respectively, according to a paired t-test with $p < 0.05$. The improvements of \textsc{WCT - cv} over other ablations are not statistically significant.}

  \vspace{-3mm}
    \label{table:ablation}
    
\end{table}

\subsection{Ablation Experiments}
\label{sec:ablation}

\paragraph{Simple fine-tuning vs. co-training.}

We perform an experiment where we first train two classifiers on $D^a$ and then continue fine-tuning them on two randomly divided subsets of $D^l$. This approach is denoted as \textsc{Simple FT - ensembled}. We compare the performance of this approach with the performance of \textsc{WCT - cv}
and show the results in Table \ref{table:ablation}. As we can see, the performance of \textsc{Simple FT - ensembled} is substantially lower than \textsc{WCT - cv} illustrating the necessity of SSL methods such as our proposed co-training method. 

\paragraph{Ensembled self-training vs. co-training.} To evaluate the necessity of co-training two classifiers by exchanging information between them, we explore a method where we also train two classifiers simultaneously but do not exchange any information. We denote this method as \textsc{WST - ensembled} which stands for \textsc{Weighted Self-training - ensembled}. Table \ref{table:ablation} shows that \textsc{WST - ensembled} obtains a lower Macro F1 than \textsc{WCT - cv}. Thus, co-training the classifiers by exchanging their knowledge is clearly more advantageous than simple ensembling.

\paragraph{Self-training with random weighting strategy vs. co-training.} To further evaluate the benefit of training two classifiers with co-training, we experiment with a weighted self-training method where we train a single classifier and randomly decide between Eq. \ref{equation:weight_assignment_1} and Eq. \ref{equation:weight_assignment_2} to assign the weight of each automatically annotated example. This self-training method is denoted as \textsc{WST - R}. As we can see in Table \ref{table:ablation}, the performance of the \textsc{WST - R} is lower than \textsc{WCT - cv}. This is because, similar to other SSL baselines, \textsc{WST - r} is more susceptible to error accumulation due to its reliance on a single classifier whereas, our co-training approach aims to learn complementary information from the data, which results in the classifiers being able to filter out different types of noise. Consequently, \textsc{WCT - cv} outperforms \textsc{WST - r}.

\paragraph{Multi-set vs. single-set co-training.} To understand the necessity of using different splits of human labeled data to train the two classifiers in our co-training approach, we explore a method named \textsc{WCT - cv Both 2K}. This method is similar to \textsc{WCT - cv} except it uses all $2,000$ human labeled examples to train both classifiers. We can see 
that the \textsc{WCT - cv Both 2K} shows a lower Macro F1 than \textsc{WCT - cv} illustrating that training the classifiers using two different splits of human labeled data enforces the classifiers to further capture complementary information.

\paragraph{High weighting vs. fine-tuning with human labeled.} 
To evaluate the necessity of utilizing automatically labeled and human labeled examples using separate steps (Step 2 and 3 of our approach, described in Section \ref{section:wct}), we perform an experiment where we combine the human labeled examples with large weights ($1.0$) with the automatically labeled examples in our co-training epochs in Step 2. This approach is denoted as \textsc{WCT - cvh}.

As we can see in Table \ref{table:ablation}, \textsc{WCT - cvh} shows a $\approx 0.4\%$ drop in performance compared with \textsc{WCT - cv} with \textsc{RoBERTa}. Inspecting the weights for the automatically annotated examples in \textsc{WCT - cvh}, we find that in the early co-training epochs, their average weight is ~0.5. This is because the classifiers in the early epochs tend to be less confident about their predictions which results in more than half of the automatically annotated examples having a weight less than 0.4. Because of the disproportionately high weights for the human labeled examples ($1.0$), the training of the classifiers become focused on them, and the automatically annotated examples remain underutilized due to their smaller weights, losing the benefits of increased data diversity. Consequently, we see a drop in the overall performance.

\vspace{-2mm}

\setlength\dashlinedash{0.2pt}
\setlength\dashlinegap{1.5pt}
\setlength\arrayrulewidth{0.3pt}
\begin{table}[t]
\centering
\small
  \begin{tabular}{l c c }
    \toprule
{\bf Approach} & {\bf ANLI R3}     & {\bf Dynasent R1} \\ 
   \midrule
   \textsc{15\% Corrupted} & $42.34$ & $78.39$ \\
   \textsc{Simple FT - ensembled} & $43.37$ & $79.47$\\
   \textsc{Co-teaching} & $43.83$ & $78.72$\\
   \textsc{WCT - cv} & $\textbf{44.43}^{*}$ & $\textbf{79.91}^{*\#}$ \\

    \bottomrule
  \end{tabular}
  \vspace{-2mm}
  \caption{Macro F1 of different methods for other tasks. * and \# indicate statistically significant improvements over \textsc{$15\%$ Corrupted} and \textsc{Co-teaching}, respectively, according to a paired t-test with $p < 0.05$. 
  }

    \label{table:other_datasets}
\end{table}

\subsection{Generalizability Analysis}
\vspace{-1mm}
\label{sec:generalizability}
To evaluate the generalizability of our proposed co-training approach, we experiment with a challenging NLI dataset from the general domain --- ANLI \cite{nie-etal-2020-adversarial} and a challenging sentiment analysis dataset --- Dynasent \cite{potts-etal-2021-dynasent}. In particular, we choose the Round 3 (R3) subset of ANLI and Round 1 (R1) subset of Dynasent. We choose these particular subsets because their training sets contain a large number of examples ($100K$ in ANLI R3, $90K$ in Dynasent R1). The experimental settings for these two datasets are described below.

We simulate a `potentially noisy' setting similar to \textsc{SciNLI} by randomly corrupting $15\%$ of the labels in the training sets of these datasets. $2,100$ non-corrupted class-balanced examples are used as $D^l$, and rest of the examples are used as $D^a$. We train \textsc{RoBERTa} models in four settings: 1) \textsc{15\% Corrupted}---the full training set after corrupting $15\%$ of the labels; 2) \textsc{Simple FT - ensembled}---same method used in Section \ref{sec:ablation}; {3) \textsc{Co-teaching}---our best performing baseline; and 4) \textsc{WCT - cv}---our proposed approach. Each experiment is run three times and the average of the Macro F1 scores are reported in Table \ref{table:other_datasets}. We find that:

\paragraph{Our co-training approach improves the performance for both ANLI and Dynasent.} 
\textsc{WCT - cv} shows better performance than \textsc{15\% Corrupted}, \textsc{Simple FT - Ensembled}, and \textsc{Co-teaching} for both \textsc{ANLI} and \textsc{Dynasent}. Therefore, while the primary objective of our proposed method is to improve the performance in scientific NLI, we can see that it can be useful for other challenging tasks.

\vspace{-1mm}
\section{Conclusion \& Future Work}
\vspace{-2mm}

In this paper, we propose a novel co-training approach for scientific NLI. In contrast to the existing approaches, we assign importance weights to automatically annotated examples based on the historical training dynamics of the classifiers. Instead of filtering out examples based on an arbitrary threshold, we decide the impact of the automatically annotated examples in subsequent training based on their assigned weights. We encourage divergence between the classifiers by assigning a contrasting weight to the \textit{ambiguous} examples to train each classifier. Our experiments show that the proposed approach obtains substantial improvements over the existing approaches. In the future, we will explore methods to further harness the training dynamics of the classifiers in improving the performance of SSL methods.

\section*{Acknowledgements}
 This research is supported by NSF CAREER award 1802358 and NSF IIS award 2107518, and UIC Discovery Partners Institute (DPI) award.  Any opinions, findings, and conclusions expressed here are those of the authors and do not necessarily reflect the views of NSF or DPI. 
We thank our anonymous reviewers for their constructive feedback, which helped improve the quality of our paper.

\section*{Limitations}
Our proposed approach shows promising performance in learning from noisy labels, but it requires a small human annotated training set. Manually annotating a small training set is significantly more feasible, and cheaper than manually removing the noisy examples from a large dataset. Nevertheless, one has to carefully consider the expense incurred for the small human annotated training set, and the obtained performance gain to employ our proposed method for other datasets and/or tasks.

Furthermore, since our approach trains two classifiers simultaneously, it requires a higher amount of computational resources. For example, we utilize two NVIDIA RTX A5000 GPUs to train the classifiers using our proposed approach whereas the single classifier based methods that we explored as our baselines are trained with a single GPU. While the necessity of higher amount of resources is a common constraint among all dual classifier based methods (e.g., \textsc{Co-teaching}), the trade-off between the performance gain with our approach and the additional computational costs should be carefully considered.

\bibliography{anthology,custom}
\bibliographystyle{acl_natbib}

\newpage
\clearpage
\appendix

\section{Details on Data Annotation}
\label{sec:annotation_details}
\vspace{-2mm}

\setlength\dashlinedash{0.2pt}
\setlength\dashlinegap{1.5pt}
\setlength\arrayrulewidth{0.3pt}
\begin{table}[t]
\centering
\small

  \begin{tabular}{l c c }
    \toprule
{\bf Class} & {\bf \#Annotated}     & {\bf Agreement} \\ 
   \midrule
    Contrasting & $624$ & $93.4\%$ \\
    Reasoning & $624$ & $82.4\%$ \\
    Entailment & $624$ & $80.1\%$ \\
    Neutral & $624$ & $94.6\%$ \\
    \hdashline
    Overall & $2,496$ & $87.6\%$\\
    \bottomrule
  \end{tabular}
  \caption{Number of manually annotated examples and the agreement rate between the gold labels and automatically assigned labels for each class.
  }
\vspace{-3mm}

    \label{table:class_wise_agreement_table}
    
\end{table}

We randomly sample a small subset of examples balanced over the classes from the full \textsc{SciNLI} training set and ask three expert annotators to annotate the label based on only the available context in the two sentences in each example. A gold label is assigned to each example based on the majority consensus of the annotated labels. If the annotators cannot reach a majority consensus, no gold label is assigned. The examples for which the human annotated gold label match with the distantly supervised automatically assigned label are selected to be in the \textbf{human annotated training set}. We iteratively continue sampling examples from the \textsc{SciNLI} training set and annotating them until we have $2000$ human annotated training examples balanced over the classes. The annotated data will be made available for academic research purposes at our GitHub link\footnote{\url{https://github.com/msadat3/weighted_cotraining}}. 

\paragraph{Agreement Rates} We iteratively sample examples from \textsc{SciNLI} training set until we had at least $500$ examples from each class in the human annotated training set. In total $2,496$ examples are annotated among which $2,187$ had a match between the gold label and the automatically assigned label based on distant supervision. We then randomly down-sample the classes with higher support to $500$ examples to balance the datasets. The annotators achieved a Fleiss-k score of $0.68$. The class-wise agreement rate can be seen in Table \ref{table:class_wise_agreement_table}. 

\paragraph{Annotator Details} Undergraduate students were hired as annotators from the authors' institution. The annotators were trained for several iterations until their annotations reached a satisfactory agreement with the authors. The trained annotators then start their final annotations for constructing the human annotated training set. The compensation was set at the hourly rate of $\$15$.

\begin{table*}[!htbp]
\centering
\small

\scalebox{0.9}{
  \begin{tabular}{l c c c c c c}
    \toprule
{\textsc{Model}} &  {\textsc{Hardware}} &  {\textsc{Networks}} & {\textsc{SWE}} & {\textsc{Security}} & {\textsc{NeurIPS}} & {\textsc{Overall}}\\

    \midrule

    \textsc{DS-labeled (101K)} & $75.60 \pm 0.8$ & $72.70 \pm 0.5$ & $74.36 \pm 0.3$ & $74.99 \pm 0.3$ & $78.35 \pm 1.0$ & $75.19 \pm 0.5$\\
    \textsc{Co-teaching} & $75.06 \pm 2.8$ & $74.03 \pm 0.7$ & $74.07 \pm 1.2$ & $74.27 \pm 1.1$ & $77.77 \pm 0.8$ & $75.04 \pm 1.2$\\
    \textsc{WCT - cc} & $77.41 \pm 0.5$ & $76.24 \pm 0.3$ & $75.68 \pm 0.4$ & $75.34 \pm 0.7$ & $79.38 \pm 0.1$ & $76.80 \pm 0.1$\\
    \textsc{WCT - cv} & $77.32 \pm 0.3$ & $76.12 \pm 0.3$ & $76.44 \pm 0.4$ & $76.46 \pm 0.7$ & $79.97 \pm 0.5$ & $77.11 \pm 0.1$\\
    \midrule

  \end{tabular}}
   \caption{\small Domain-wise and overall Macro F1 scores (\%) of \textsc{DS-labeled (101K)} and \textsc{Co-teaching} baselines compared with our methods \textsc{WCT - cc} and \textsc{WCT - cv} trained using \textsc{SciNLI} and tested on \textsc{MSciNLI}. 
  }
  \vspace{2mm}
  
    \label{table:ood_performance}
\end{table*}

\setlength\dashlinedash{0.2pt}
\setlength\dashlinegap{1.5pt}
\setlength\arrayrulewidth{0.3pt}
\begin{table*}[!htbp]
\centering
\small

  \begin{tabular}{p{40em}}
    \toprule
  <human>: Consider the following two sentences:\newline
    Sentence1: <sentence1>\newline
Sentence2: <sentence2>\newline
Based on only the information available in these two sentences, which of the following options is true?\newline
a. Sentence1 generalizes, specifies or has an equivalent meaning with Sentence2.\newline
b. Sentence1 presents the reason, cause, or condition for the result or conclusion made Sentence2.\newline
c. Sentence2 mentions a comparison, criticism, juxtaposition, or a limitation of something said in Sentence1.\newline
"d. Sentence1 and Sentence2 are independent.\newline
<bot>:\\
    \bottomrule
  \end{tabular}

  \caption{Prompt template used for our experiments with LLM. Here, <X> indicates a placeholder X which is replaced in the actual prompt. }

    \label{table:prompts}
    \vspace{2mm}
\end{table*}

\begin{table*}[!htbp]
\centering
\small
\scalebox{1.00}{
  \begin{tabular}{l c c c c c}
    \toprule
{\textsc{Prompt Type}} & {\textsc{Contrasting}} &  {\textsc{Reasoning}} & {\textsc{Entailment}} & {\textsc{Neutral}} & {\textsc{Macro F1}}\\

    \midrule
    \textsc{Zero-shot} & $45.55$ & $37.72$ & $15.61$ & $18.50$ & $29.34$ \\
    \textsc{Few-shot} & $57.70$ & $37.18$ & $51.51$ & $63.90$ & $52.57$ \\
   \midrule
  \end{tabular}}
   \caption{Class-wise F1 and overall Macro F1 of Zero-shot and Few-shot prompting with Llama-2-13b-chat for \textsc{SciNLI} test set. 
  }
  
    \label{table:LLM_results}
\end{table*}

\section{Implementation Details}
\label{sec:implementation}

We use the `scibert-scivocab-cased' and the `roberta-base' variants of \textsc{SciBERT} and \textsc{RoBERTa}, respectively for all our baselines and proposed methods. We use the huggingface\footnote{\url{https://huggingface.co/docs/transformers/index}} implementation for both models. The premise and hypothesis for each example are concatenated with a \texttt{[SEP]} token and the \texttt{[CLS]} token's representation is sent through a fully connected layer with softmax activation to predict the class. For all methods (baselines and proposed), the batch size is set at 64. We use an Adam optimizer \cite{kingma2014adam} to train the models and set the learning rate at $2e-5$. We use a learning rate of $2e-6$ for fine-tuning the models in the last step of our proposed approach (Step 3 in Section \ref{section:wct}). For backtranslation, we use a transformer based sequence-to-sequence model\footnote{\url{https://huggingface.co/Helsinki-NLP/opus-mt-en-ROMANCE}}. The confidence threshold for \textsc{DBST}, \textsc{FixMatch}, and the base threshold for \textsc{FlexMatch} is set at $0.9$. For the consistency regularization based algorithms such as \textsc{FixMatch}, \textsc{FlexMatch}, and \textsc{SoftMatch}, we use a ratio of $1:7$ between labeled and unlabeled data. The weight for the supervised loss for the consistency regularization methods is set at 0.8, and that of the unsupervised loss is set at 0.2. For both \textsc{Co-teaching} and \textsc{Co-teaching+}, we set the estimated noise rate, $\epsilon = 0.15$. 
For the \textsc{DS}, \textsc{BT}, \textsc{FS}, and the consistency regularization baselines, we train the classifiers for a maximum of $10$ epochs. For the other experiments, we set the number of maximum epochs at $5$. Early stopping is employed with a patience of 2. Macro F1 score on the development set is used as the stopping criteria for early stopping.

We use a single NVIDIA RTX A5000 GPU for the experiments involving a single model training (e.g., DBST, FixMatch etc.), and two NVIDIA RTX A5000 GPU for the dual model training experiments (e.g., co-training, weighted co-training, co-teaching). Each co-training experiment takes $\approx 5$ hours to complete.

\section{Robustness Analysis}
\label{appendix:robustness}

We assess the robustness of our proposed approach by experimenting with out-of-domain (OOD) test sets from \textsc{MSciNLI} \cite{sadat-caragea-2024-mscinli} that covers the following domains from computer science: ``Hardware'', ``Networks'', ``Software \& its Engineering'', ``Security \& Privacy'', and ``NeurIPS'' which is related to machine learning. Particularly, we evaluate the following models that are trained using the training set of \textsc{SciNLI} (that covers computational linguistics only) on the \textsc{MSciNLI} test set: \textsc{DS-labeled (101K)}, \textsc{Co-teaching}, \textsc{WCT - cc}, and \textsc{WCT - cv}. The domain-wise and overall Macro F1 of these methods for \textsc{MSciNLI} are reported in Table \ref{table:ood_performance}. We find the following:

\paragraph{Our proposed approach is robust in an OOD setting.} As we can see, for most of the domains in \textsc{MSciNLI}, \textsc{Co-teaching} fails to show any improvement over \textsc{DS-labeled (101K)}. In contrast, \textsc{WCT - cc} outperforms \textsc{DS-labeled (101K)} in all domains resulting in an overall increase of $1.61\%$. In addition, \textsc{WCT - cv} improves the OOD performance further over \textsc{WCT - cc} and shows an improvement of  $1.92\%$ in Macro F1 over the \textsc{DS-labeled (101K)} baseline. These results illustrate that, our proposed approach not only improves in-domain performance, but also trains robust models that perform well for OOD datasets.

\section{LLMs for SciNLI}
\label{appendix:LLM_results}
We experiment with \textit{Llama-2-13b-chat-hf} \cite{touvron2023llama} for \textsc{SciNLI}. We design a prompt template that first presents the two sentences from each example, and asks a multiple choice question with the class definitions as the choices. The prompt template can be seen in Table \ref{table:prompts}. Note that we omitted the special tags needed to prompt the LLM (e.g., \texttt{[INST]}) in the Table for brevity. We perform experiments with the LLM in 2 settings:
\begin{itemize}
    \item Zero-shot: no exemplars are shown to the LLM.
    \item Few-shot: one exemplar from each class is prepended to the prompt.
\end{itemize}
The results from these experiments can be seen in Table \ref{table:LLM_results}. As we can see, the \textsc{Few-shot} performance is substantially higher than the \textsc{Zero-shot} performance. Nevertheless, the Macro F1 of even the \textsc{Few-shot} setting is only $52.57\%$ illustrating that making use of LLMs for reducing the noise in the \textsc{SciNLI} trainign set is not viable approach.

\end{document}